\documentclass[letterpaper, 10 pt, conference]{ieeeconf}  

\IEEEoverridecommandlockouts                              

\usepackage{graphics} 
\usepackage{epsfig} 
\usepackage{mathptmx} 
\usepackage{times} 
\usepackage{amsmath} 
\usepackage{amssymb}  

\usepackage{bbding}
\usepackage{gensymb}
\usepackage[utf8]{inputenc} 
\usepackage[T1]{fontenc}    
\usepackage{hyperref}       
\usepackage{url}            
\usepackage{booktabs}       
\usepackage{amsfonts}       
\usepackage{nicefrac}       
\usepackage{microtype}      
\usepackage{color}
\usepackage[table]{xcolor}
\usepackage{cleveref}
\usepackage{graphicx}
\usepackage{float}
\usepackage{multirow}
\usepackage{diagbox}

\usepackage{algorithm}
\usepackage{algpseudocode}
\usepackage{arydshln}
\usepackage{colortbl}

\usepackage{multicol}
\usepackage{multirow}
\usepackage{listings}
\definecolor{dkgreen}{rgb}{0,0.6,0}
\definecolor{gray}{rgb}{0.5,0.5,0.5}
\definecolor{mauve}{rgb}{0.58,0,0.82}
\newcommand{\rvlnbert}{VLN $\protect\circlearrowright$ BERT}
   
\title{\LARGE \bf
Discuss Before Moving: Visual Language Navigation \\via Multi-expert Discussions
}

\author{Yuxing Long, Xiaoqi Li, Wenzhe Cai, Hao Dong
\thanks{All authors are with Hyperplane Lab, CFCS, School of CS, Peking University and National Key Laboratory for Multimedia Information Processing. Xiaoqi Li and Hao Dong are also with Beijing Academy of Artificial Intelligence (BAAI). Wenzhe Cai is also with the School of Automation, Southeast University.}
\thanks{
Corresponding to hao.dong@pku.edu.cn}
}

\begin{document}   
\maketitle

\begin{abstract}  
Visual language navigation (VLN) is an embodied task demanding a wide range of skills encompassing understanding, perception, and planning. For such a multifaceted challenge, previous VLN methods totally rely on one model's own thinking to make predictions within one round. However, existing models, even the most advanced large language model GPT4, still struggle with dealing with multiple tasks by single-round self-thinking. In this work, drawing inspiration from the expert consultation meeting, we introduce a novel zero-shot VLN framework. Within this framework, large models possessing distinct abilities are served as domain experts. Our proposed navigation agent, namely DiscussNav, can actively discuss with these experts to collect essential information before moving at every step. These discussions cover critical navigation subtasks like instruction understanding, environment perception, and completion estimation. Through comprehensive experiments, we demonstrate that discussions with domain experts can effectively facilitate navigation by perceiving instruction-relevant information, correcting inadvertent errors, and sifting through in-consistent movement decisions. The performances on the representative VLN task R2R show that our method surpasses the leading zero-shot VLN model by a large margin on all metrics. Additionally, real-robot experiments display the obvious advantages of our method over single-round self-thinking.

\end{abstract}

\section{INTRODUCTION}
Visual language navigation (VLN) is an exciting and emerging research field focused on developing embodied agents capable of following natural language instructions to navigate real 3D environments. Achieving success in VLN is essential for the creation of service robots designed for personal and domestic use, which has the potential to significantly impact human life. Therefore, visual language navigation has become a research hotspot within the field of embodied AI~\cite{anderson2018vision,krantz2020beyond,he2021landmark,Ku2020RxR}, leading to substantial advancements in both academic and industrial domains.

\begin{figure}[t]
\begin{center}
    \includegraphics[width=0.85\linewidth]{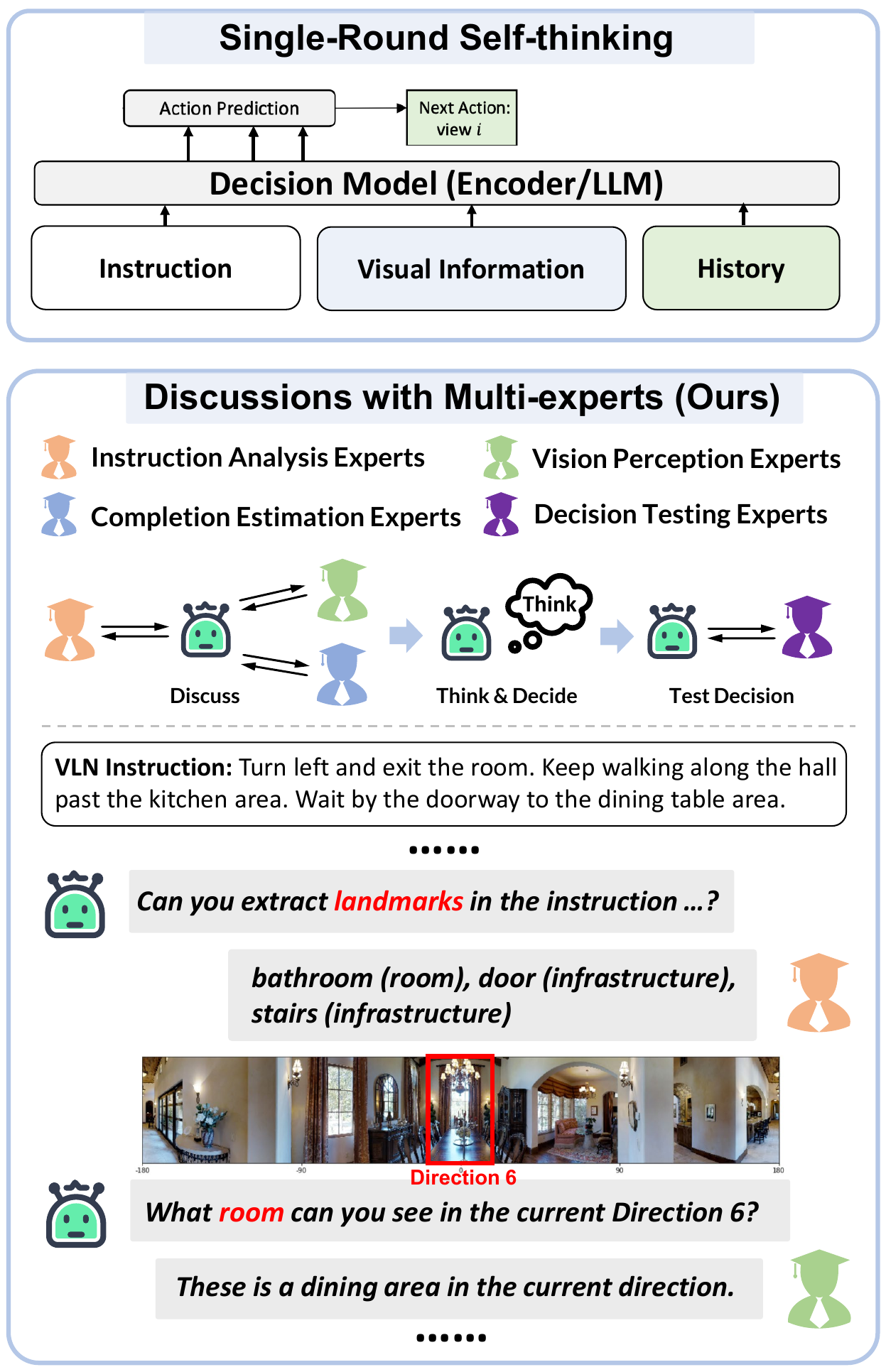}
\end{center} 
\vspace{-0.4cm}
\caption{Comparison between single round self-thinking and multi-expert discussions inference paradigms. The single-round self-thinking models passively take all accessible navigation information as input and have to make a prediction in one execution, while our DiscussNav agent relieves from completing complex reasoning at one time and can actively obtain the needed information via multi-expert discussions.}
\label{fig:teasor}
\vspace{-0.5cm}
\end{figure}

Following instructions to navigate the indoor scene is a very challenging embodied task, demanding keen observation and meticulous thinking. At each action step, the navigation agent needs to understand all actions and landmarks specified in the instruction, observe the panoramic scene to identify crucial objects, and review the accessed position to estimate the instruction completion. Every movement decision relies on the successful execution of these sub-tasks, each of which demands different knowledge and abilities. 

Facing such a multi-faceted challenge, previous methods totally rely on one decision model, either specially trained VLN model~\cite{anderson2018vision,fried2018speaker,tan2019learning,hao2020towards,hong2021vln, chen2021history,chen2022think} or zero-shot large language model~\cite{navgpt}, to make every step movement prediction. As shown at the top of Figure~\ref{fig:teasor}, these models passively take all accessible navigation information as input and directly make a movement prediction by their own knowledge and reasoning within one execution. This kind of inference is a single-round self-thinking paradigm, which requires the decision model to independently deal with multiple VLN subtasks simultaneously based on the pre-prepared information. However, even the most advanced large language model GPT4 still struggles with completing several different domain tasks at the same time. While training models with VLN simulator data can enhance their performances within simulated environments, the substantial sim-to-real gap in camera setting and environment difference poses a significant barrier to their deployment on the real robot. As shown in Table~\ref{tab:real}, the trained model with more than 70\%  simulation success rate cannot complete one instruction in the real world, which indicates that it does not truly have the ability to cope with multiple navigation subtasks via current single-round self-thinking. Therefore, we explore a new inference paradigm in this work to address problems of single-round self-thinking.

In human society, international organizations or national governments frequently convene expert consultation meetings~\cite{2006Breast, world2021meeting, stojkov2018hot, world2017expert} to address issues within specialized domains such as health, environment, and energy. These gatherings serve as platforms for in-depth discussions with domain experts, providing valuable insights and information essential for informed decision-making. The expert consultation meeting brings us inspiration. If the navigation agent can discuss with multiple domain experts before moving, they will have more confidence to make movement decisions. 

Recent advancements in large-scale training have empowered large models with various domain knowledge and abilities. 
We discover that their potential navigation capabilities can be activated by special prompts, which sparks our idea of creating domain experts with large models. Therefore, we analyze the reasoning process about VLN and abstract four crucial subtasks including instruction analysis, vision perception, completion estimation, and decision testing. Following subtask requirements, we craft on assigning the expert's role and defining the corresponding task in a series of prompts. Eight different domain experts for VLN are created by instructing large models with these prompts. With these experts, we design a zero-shot VLN agent driven by the large language model. Rather than completely think by itself, the DiscussNav can actively discuss with multi-experts before making a movement decision as shown at the bottom of Figure~\ref{fig:teasor}. Compared with single-round self-thinking, the multi-expert discussion can effectively reduce the burden on the decision model and achieve better performance through the collective power of multiple large models. Besides, in the discussion process, we incorporate a verification mechanism for action decomposition and completion estimation to rectify errors that might arise during the discussion.

In this work, our main contributions are:
\begin{itemize}
    \item[$\bullet$] We introduce DiscussNav, a novel zero-shot navigation agent that leverages collective power through active discussions with multiple domain experts.
    \item[$\bullet$] We create multiple domain experts for visual language navigation by assigning specific roles and tasks to large models, completely eliminating human involvement.
    \item[$\bullet$] Experiments on the representative visual language navigation task R2R and real-robot show the effectiveness of discussions with domain experts on the navigation agent's zero-shot performance.
\end{itemize}

\begin{figure*}[t]
\begin{center}
    \includegraphics[width=1\linewidth]{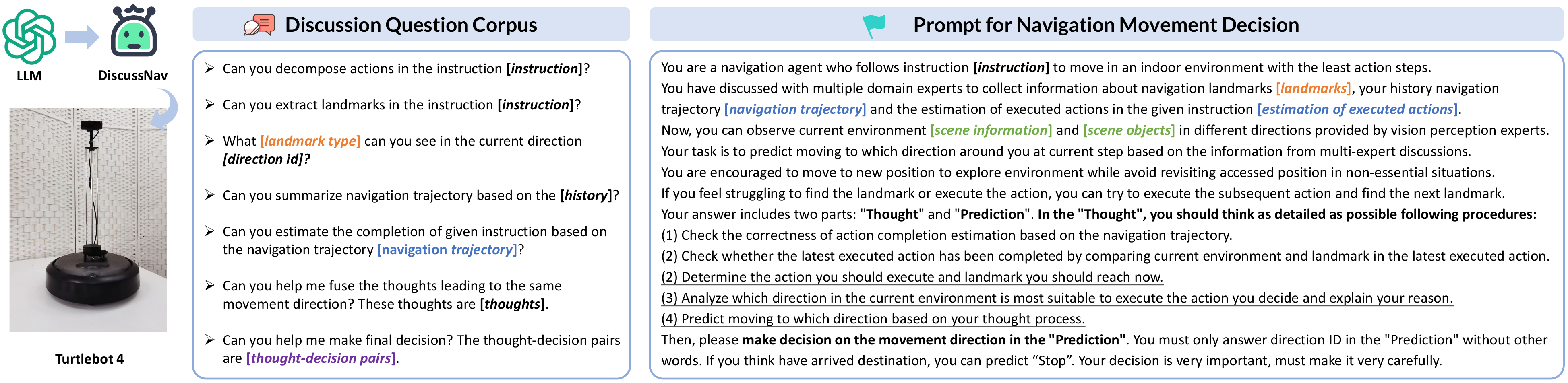}
\end{center}
\vspace{-0.4cm}
\caption{Demonstration of navigation agent DiscussNav powered by large language model (GPT4). The DiscussNav agent fills in question templates in the corpus to discuss with multiple domain experts and prompts the large language model to make navigation decisions based on the multi-expert discussion results. The ``[\textbf{\textit{xxx}}]'' are slots that should be filled with specified information. We use color~(See Fig~\ref{fig:method}) to distinguish different expert discussion results. In the Matterport3D simulation~\cite{mattersim} test, the DiscussNav agent will choose the first candidate viewpoint of the predicted direction to move.}
\label{fig:discussnav}
\vspace{-0.4cm}
\end{figure*}
    
\section{RELATED WORK}
\subsection{Vision-and-Language Navigation}
Language-driven vision navigation is demanded by widely applicable embodied navigation agents. Previous supervised methods make efforts on 
data augmentation~\cite{liu2021vision,wang2022less,li2022envedit,tan2019learning,parvaneh2020counterfactual,fu2020counterfactual,wang2022counterfactual}, memory mechanisms~\cite{chen2021history,wang2021structured,pashevich2021episodic}, 
and pre-training~\cite{majumdar2020improving,hao2020towards,guhur2021airbert,wu2022cross,qi2021road}. These methods all focus on how to effectively utilize simulator data in the training to improve simulated performance but ignore the real-world deployment. Although some researchers~\cite{cvdn, jusask, weta, teach} try to equip navigation agents with dialogue abilities, they all involve human annotation or participation. Recently, the development of large models facilitated research on zero-shot visual language navigation agents that can navigate in the real world. LM-Nav~\cite{shah2023lm} employs GPT-3~\cite{brown2020language} in an attempt to identify landmarks or subgoals on the predefined graph, while CoW~\cite{gadre2022cow} and VLMaps~\cite{huang2022visual} performs zero-shot language-based visual navigation by creating maps with multimodal models in the prior exploration. Being different from these methods dependent on predefined graphs or prior exploration, NavGPT~\cite{navgpt} builds an out-of-the-box visual language navigation framework that includes visual scene semantics into prompts for the LLM to directly perform VLN. However, all the above methods follow the single-round self-thinking paradigm, which limits their performance on VLN task and hinders their deployments on real robots. Our work focuses on leveraging the collective power of multiple large models relived in discussions to construct a more advanced out-of-the-box visual language navigation framework.

\subsection{Large Models}
With the massive success in large-scale model training, a new cohort of large models has shown evolutionary progress toward achieving Artificial General Intelligence (AGI). Large language models (LLMs)~\cite{gpt3,chatgpt,openai2023gpt4} have equipped with a wide range of impressive abilities in information summary, code generation, task planning, and so on. In the recent, a series of research has been conducted on the LLMs to explore how to further enhance LLMs by in-context learning examples, chain-of-thought prompting~\cite{wei2022chain,kojima2022large,gao2022pal,Wang2023Self}, and external tool use~\cite{schick2023toolformer,shen2023hugginggpt}. Drawing inspiration from instruction tuning within LLMs, researchers in the multimodal domain have taken strides toward integrating visual instruction tuning to enhance the instruction-following ability of multimodal large models. Multimodal large models, such as LLaVA~\cite{liu2023visual}, VisualGLM~\cite{du2022glm,ding2021cogview}, and InstructBLIP~\cite{instructblip}, display remarkable capabilities in responding to instructions or questions regarding image content, nuanced semantics, and image-to-text artistic creation. These endeavors lay the groundwork for the development of embodied agents based on large models.


\section{METHOD}
Inspired by human behavior, we design a zero-shot visual language navigation framework. The navigation agent DiscussNav can actively initiate discussions with domain experts driven by large models to collect needed information before the agent makes movement decisions. In this section, we will introduce how to form the DiscussNav agent and domain experts. Then, we will illustrate how DiscussNav conducts navigation discussions with experts before every step moving.

\subsection{DiscussNav Agent for Visual Language Navigation}
We create the DiscussNav agent based on the discussion question corpus and the powerful large language model GPT4. The left box in Fig~\ref{fig:discussnav} shows the discussion question corpus. In the navigation process, the DiscussNav agent will one-by-one fill in the slots of these question templates and discuss them with corresponding domain experts to collect information for navigation. The right box in Figure~\ref{fig:discussnav} displays how the DiscussNav agent instructs the large language model to make movement decisions according to the multi-expert discussion results about landmarks, navigation trajectory, competition estimation, and scene visual information. The large language model GPT4 is required to first conduct a chain-of-thought following a series of requirements including checking the correctness of completion estimation and then write down a movement decision in the ``Prediction'' field for the convenience of resolution. At each inference time, following the beam search mechanism, the large language model can produce $N$ different responses simultaneously for one input. Compared with the greedy search, beam search can cover more possibilities. The DiscussNav agent will discuss these predictions with decision testing experts for the final movement decision.

\subsection{Domain Experts Driven by Large Models}
\begin{figure*}[t]
\begin{center}
    \includegraphics[width=1\linewidth]{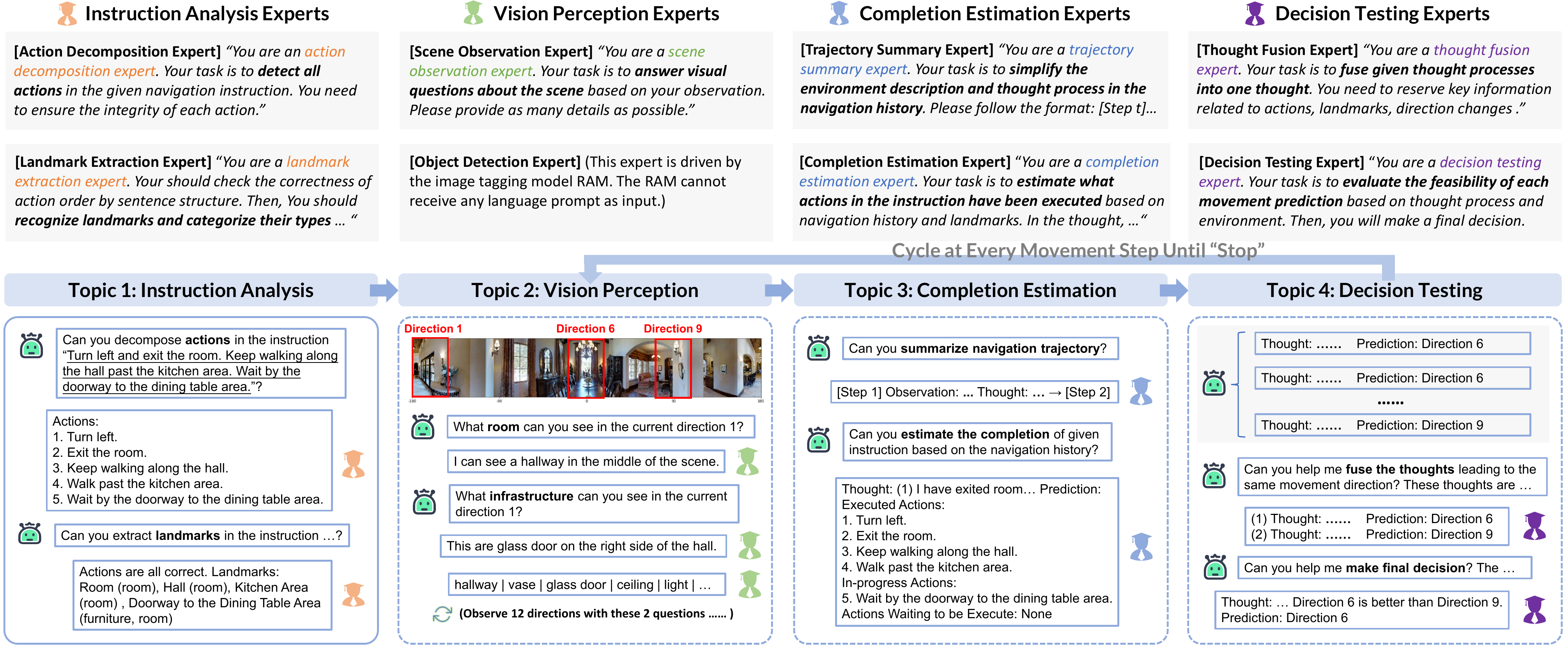}
\end{center}
\vspace{-0.4cm}
\caption{Establishments of domain experts and discussions between the DiscussNav agent and multi-experts. After receiving the VLN instruction, DiscussNav will first discuss with instruction analysis experts to learn about actions and landmarks. Then, at every movement step, DiscussNav will communicate with vision perception experts about landmark-relevant visual information in surrounding directions and interact with completion estimation experts about executed actions. Based on discussion results, DiscussNav will make $N$ different predictions and invite decision testing experts to decide final movement direction.}
\label{fig:method}
\vspace{-0.4cm}
\end{figure*}

Large-scale training on multi-source datasets equips large models with rich domain knowledge and abilities. Meanwhile, the auto-regressive instruction tuning makes both large language models and multimodal large models sensitive to the given prompt. Utilizing this feature, we can explore large models' professional knowledge and activate their domain capabilities by assigning specific roles and defining clear task requirements in the inputted prompt. Therefore, as the top row of Figure~\ref{fig:method} shows, we establish multiple domain experts for visual language navigation through specific prompts. The creation process will be introduced in the following. 

\textbf{Instruction Analysis Experts.}
The navigation instruction usually involves a series of actions and landmarks. Some actions and landmarks are not easy to understand due to grammatical structures. Take the instruction \textit{``Stop just past the eye exam chart on the wall''} as an example, the action sequences are \textit{``walk past''}, \textit{``stop''} rather than the reverse order while the landmark is \textit{``the eye exam chart on the wall''} instead of \textit{``eye exam chart''}, \textit{``wall''}.
To accurately identify action sequences, we allocate \textbf{action decomposition expert}'s role to GPT4 and define the task of detecting all actions in the given navigation instruction. The expert takes raw human instruction as input and is expected to produce decomposed actions in a sequential manner. Furthermore, we create GPT4 driven \textbf{landmark extraction expert} with the goal of recognizing landmarks and categorizing their types. We impose the requirement to ``ensure the integrity of each landmark'' to facilitate the expert. Note that, to guarantee the correctness of action decomposition, we add an error correction task before the landmark expert's formal task, which requires the expert to check the action order based on the sentence structure of the original instruction.

\textbf{Vision Perception Experts.}
Observing instruction landmarks existing in the surrounding environment is essential for the navigation agent to make decisions on the next movement direction. The instruction landmarks can be roughly divided into scene-level landmarks, such as \textit{``bedroom''}, \textit{``kitchen''} and \textit{``living room''} and object-level landmarks like \textit{``stair''}, \textit{``sink''} and \textit{``couch''}. We create \textbf{scene observation expert} by multimodal large model InstructBLIP~\cite{instructblip}, which takes scene image and scene queries as input and is expected to observe required scene-level visual information. In the navigation discussion, the expert can observe specific scene-level information such as room type, infrastructure, and furniture following the task requirement. The \textbf{object detection expert} is built upon the Recognize Anything Model (RAM)~\cite{ram}. Due to the limitation of RAM's architecture, this expert can only take one image as input and predict object tags. Compared with the scene observation expert, the object detection expert has the advantage of discovering inconspicuous objects in the scene, which can find most existing objects in the scene.

\textbf{Completion Estimation Experts.}
In the navigation process, each action may require more than one step to execute, which leads to the misalignment between the movement steps and executed actions. The navigation agent should be aware of what actions they have completed and what next action they need to execute in the current environment. Firstly, we allocate \textbf{trajectory summary expert}'s role to ChatGPT and define its task of simplifying the environment description and thought process in the navigation history. Taking navigation history as input, the expert will remove redundant objects in the environment description and filter non-essential reasoning in the thought process. To make the trajectory clear, the output of this expert follows the format ``[$Step\;\;t$] $Observation: ... Thought: ...$''. Then, we create \textbf{completion estimation expert} by GPT4. The expert is required to first establish a chain-of-thought process to estimate what actions have been executed based on the historical trajectory. In the thinking, the expert needs to (1) check the passed landmarks in the navigation trajectory, (2) analyze the direction change at each movement step, and (3) estimate what actions have been executed. Such a thinking process can further improve the accuracy of expert estimation. Then, the agent should write down ``Executed Actions'', ``In-progress Actions'' and ``Actions Waiting to be Executed'' in the ``Prediction'' field.

\textbf{Decision Testing Experts.}
With the beam search setting, the navigation agent can output $N$ diverse predictions at each time. If there are differences between these $N$ predictions, the agent has to select one prediction to execute. To tackle this challenge, we first allocate \textbf{thought fusion expert}'s role to ChatGPT and define its task to ``fuse given thought processes into one thought''. The expert will take thought processes that lead to the same movement decision as input and summarize their reasons into one. Then, we require GPT4 to assume the role of \textbf{decision testing expert} and complete the tasks including ``evaluate the feasibility of each movement prediction based on thought process and current environment'' and ``select the most reliable prediction as the final decision''. The expert will read all thought-prediction pairs at the current step to compare their reasoning process following the chain-of-thought thinking and select the most suitable prediction in the "Prediction" field as the navigation agent's final movement decision at the current step.

\begin{figure*}[t]
\begin{center}
    \includegraphics[width=\linewidth]{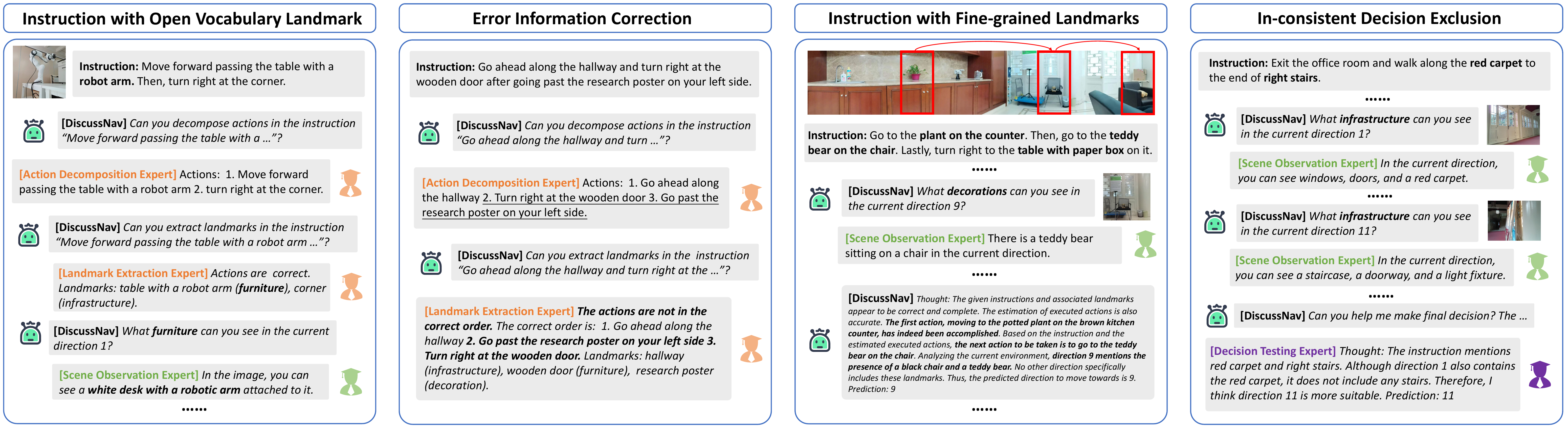}
\end{center}
\vspace{-0.4cm}
\caption{The qualitative results of DiscussNav's performance on the real robot. Through discussions with domain experts, DiscussNav can observe open vocabulary landmarks and navigate to fine-grained landmarks. In the discussion, error information can be timely corrected by other experts or the DiscussNav agent, which reduces the error accumulation. Besides, in-consistent movement decisions can be filtered by discussions with decision testing experts.}
\label{fig:case}
\vspace{-0.4cm}
\end{figure*}

\subsection{Navigation Discussions with Multiple Experts}
With the navigation agent DiscussNav and domain experts, we define a discussion mechanism stipulating the discussion order and objects. In this part, we will introduce how our DiscussNav agent conducts discussions with multiple experts.

\textbf{Instruction Analysis.}
After receiving a navigation instruction from a human, DiscussNav initially consults the action decomposition expert ``Can you decompose actions in the instruction \{instruction\}?''. The expert will break the navigation instructions into a sequence of actions. Then, DiscussNav asks the landmark extraction expert ``Can you extract landmarks in the instruction \{instruction\}?''. The expert will reply with navigational landmarks and corresponding types. Through these interactions, DiscussNav gains a comprehensive understanding of the actions it should execute and the landmarks it must pass during the navigation process.

\textbf{Vision Perception.}
At each movement step, the DiscussNav will interact with the vision perception experts to update its cognition of the surrounding environment. In this work, we set the field of view in each direction as $30\degree$, which are 12 directions in total from $0\degree$ to $360\degree$ for every position. When discussing scene-level visual information, to reduce hallucinations of the scene observation expert, the agent will formulate questions based on the landmark types rather than directly ask about the existence of concrete landmarks. For instance, if the landmark is associated with a room, such as ``kitchen'', the agent will prompt the expert ``What room can you see in the current direction \{direction id\}''.  In cases involving landmarks defined by color like ``red carpet'', the DiscussNav will ask the expert to ``What color of objects can you see in the current direction \{direction id\}''. These targeted discussions help the agent concentrate on visual information relevant to the instruction. Leveraging the RAM's feature, the communication between DiscussNav and the object detection expert is direct. DiscussNav simply presents the current environment view to the expert and receives labels for all objects in the scene as feedback.

\textbf{Completion Estimation.}
After perceiving the environment, the navigation agent should determine which action it needs to execute. To collect more information for the decision, the agent will communicate with the completion estimation experts. To begin, DiscussNav provides the navigation history, encompassing both past observations and thought processes, to the trajectory summary expert and instructs it to filter out unrelated information and summarize each step trajectory. Subsequently, the agent presents the decomposed actions and formatted trajectory to the completion estimation expert, learning about "Executed Actions", "In-progress Actions" and "Actions Waiting to be Executed". The history trajectory and feedback from experts will guide the movement planning. 

\textbf{Movement Decision.}
Through multi-round discussions with different domain experts, the agent acquires comprehensive information about instruction requirements, visual information in 12 surrounding directions, and the ongoing execution state. With this information, the agent will conduct independent chain-of-thoughts to make $N$ predictions. If these predictions lead in the same direction, the DiscussNav agent will directly execute it to move. However, if there is inconsistency among these $N$ predictions, the DiscussNav agent will turn to decision testing experts for help.

\textbf{Decision Testing.}
To make the final decision from $N$ predictions, the agent will deliver these thought-prediction pairs to the decision testing experts. The thought fusion expert first classifies thought processes leading to the same movement prediction as one group and fuses these thought groups respectively. To illustrate, for five thought-prediction pairs, if there are three pairs corresponding to position $A$ while the remaining two point to position $B$, the expert will conclude the $A$-position thought-prediction pair and $B$-position pair. Consequently, the decision testing expert will analyze these thought-prediction pairs based on the current environment to formulate a final decision for DiscussNav' movement.

\section{EXPERIMENTAL}
\subsection{Experiment Setup}
\textbf{Implementation Details}
For the development of domain experts, We construct two instruction analysis experts, a completion estimation expert, and a decision testing expert by GPT4~\cite{openai2023gpt4} with 0 temperature and 1 output (\emph{i.e.} $n=1$) setting. Additionally, we build a trajectory summary expert and a thought fusion expert by ChatGPT~\cite{chatgpt} employing the same hyperparameter configuration. 
The scene observation expert is supported by InstructBLIP $\text{FlanT5}_\text{XL}$~\cite{instructblip} and the backbone of the object perception expert is RAM-14M~\cite{ram}. For the DiscussNav agent, we drive it by GPT4 with 1 temperature and 5 output (\emph{i.e.} $n=5$) setting. Our simulation test is conducted in the Matterport3D simulator~\cite{mattersim} and our real-robot test is deployed on the \textit{Turtlebot 4 Lite}.

\textbf{Evaluation Metrics}
Our evaluation follows standardized metrics from the R2R dataset. These include Trajectory Length (TL), denoting average path length in meters; Navigation Error (NE), representing the average distance in meters between the agent’s final location and the target; Success Rate (SR), indicating the proportion of paths with NE less than 3-meter; Oracle Success Rate (OSR), SR given oracle stop policy; and SR penalized by Path Length (SPL).

\subsection{Simulator Quantitative Experiments}
\textbf{Comparison with Previous Methods.}
The R2R is the representative visual language navigation dataset. Following~\cite{navgpt} setting, We apply the 783 trajectories in the 11 val unseen environments for comparison to previous approaches. We compare our DiscussNav with previous navigation models evaluated R2R task. According to their training schema, previous methods can be categorized into three types. ``Train Only'' methods are directly trained on the R2R train dataset while ``Pretrain + Finetune'' methods are pretrained on multiple specially designed proxy tasks before being finetuned on the R2R downstream task. For ``Zero-shot'' methods, they explore completing R2R task without any navigation training or prior exploration. As shown in the Table~\ref{tab:main}, we can observe that our DiscussNav outperforms all ``Zero-shot'' methods and two ``Train Only'' methods on all five metrics by a large margin. Compared with NavGPT, our method achieves 26.47\% Success Rate improvement and 37.93\% SPL improvement. 

\definecolor{Gray}{gray}{0.9}
\begin{table}[ht]
\small
\centering
\caption{Results on R2R validation unseen split.}
\label{tab:main} 
\resizebox{1\linewidth}{!}
{
\begin{tabular}{llccc>{\columncolor{Gray}}c>{\columncolor{Gray}}c}
\toprule
\multicolumn{1}{c}{Training Schema} & \multicolumn{1}{c}{Method} &
\multicolumn{1}{c}{TL} & \multicolumn{1}{c}{NE$\downarrow$} & \multicolumn{1}{c}{OSR$\uparrow$} & \multicolumn{1}{c}{SR$\uparrow$} & \multicolumn{1}{c}{SPL$\uparrow$} \\
\midrule
\multirow{3}{*}{Train Only} 
& Seq2Seq~\cite{anderson2018vision}       & 8.39  & 7.81 & 28 & 21 & - \\
& Speaker Follower~\cite{fried2018speaker} & - & 6.62 & 45 & 35 & - \\
& EnvDrop~\cite{tan2019learning} & 10.70 & 5.22 & - & 52 & 48 \\
\midrule    
\multirow{4}{*}{Pretrain + Finetune} 
& PREVALENT~\cite{hao2020towards} & 10.19 & 4.71 & - & 58 & 53 \\
& \rvlnbert~\cite{hong2021vln}    & 12.01 & 3.93 & 69 & 63 & 57 \\
& HAMT~\cite{chen2021history}      & 11.46 & 2.29 & 73 & 66 & 61 \\
& DuET~\cite{chen2022think}      & 13.94 & 3.31 & 81 & 72 & 60 \\
\midrule
\multirow{3}{*}{Zero-shot} 
& DuET (Init. LXMERT~\cite{tan2019lxmert}) & 22.03 & 9.74 & 7 & 1 & 0 \\
& NavGPT~\cite{navgpt} & 11.45 & 6.46 & 42 & 34 & 29 \\
& DiscussNav (Ours) &9.69  &5.32  &61  &43  &40  \\
\bottomrule
\end{tabular}
}
\end{table}

\textbf{The effect of Discussions with Domain Experts.}
To investigate the effectiveness of discussions with domain experts, we conduct additional experiments to ablate different experts. We follow the previous non-train work~\cite{navgpt} to construct a new validation split sampling both from the original training set and the validation unseen set for the evaluation of zero-shot ability in various environments. The scenes from the training and validation unseen set are 61 and 11 respectively. We randomly picked 1 trajectory from these 72 environments to conduct the ablation study. As Table~\ref{tab:ablation} displays, when we ablate discussions with different experts, the DiscussNav agent's performance shows varying declines. The ablation study can prove the effectiveness of multi-expert discussions in facilitating VLN task.

\definecolor{Gray}{gray}{0.9}
\begin{table}[ht]
\small
\centering
\caption{Ablation Study about Discussions with Experts.}
\label{tab:ablation}
\scalebox{0.74}
{
\begin{tabular}{lccc>{\columncolor{Gray}}c>{\columncolor{Gray}}c}
\toprule
\multicolumn{1}{c}{Method} &
\multicolumn{1}{c}{TL} & \multicolumn{1}{c}{NE$\downarrow$} & \multicolumn{1}{c}{OSR$\uparrow$} & \multicolumn{1}{c}{SR$\uparrow$} & \multicolumn{1}{c}{SPL$\uparrow$} \\
\midrule
DiscussNav &\textbf{9.69}  &\textbf{5.32}  &\textbf{61}  &\textbf{43}  &\textbf{40}  \\
\quad w/o Instruction Analysis Experts &8.36  &5.85  &52.04  &38.77   &35.04  \\
\quad w/o Vision Perception Experts &8.41  &5.87  &46.75  &32.46   &29.01  \\
\quad w/o Completion Estimation Experts &8.21  &6.64  &42.86  &32.65  &30.13  \\
\quad w/o Decision Testing Experts &8.89  &6.30  &50.96  &37.50  &33.31  \\
\bottomrule
\end{tabular}
}
\end{table}

\subsection{Real Robot Quantitative Experiments}
We perform real robot experiments based on the \textit{Turtlebot 4 Lite} mobile robot. The \textit{Turtlebot 4 Lite} is equipped with an OAK-D Lite camera and an RPLIDAR-A1 lidar. OAK-D Lite Camera is mounted at a 60 cm height level while the lidar is only used for local obstacle avoidance without any mapping. We deploy state-of-the-art pretraining model~\cite{chen2022think}, zero-shot model~\cite{navgpt}, and DiscussNav on the Turtlebot to navigate in a semantically rich house. At every movement step, the robot will move 0.5m toward the predicted direction if there are no obstacles within one meter. Otherwise, it will move half the distance from the obstacle. For testing purposes, we define 20 different navigation instructions, which includes open-vocabulary landmark like ``robot arm'', fine-grained landmarks, and multiple room changes. 

The quantitative results of real robot experiments are displayed in Table~\ref{tab:real}. From the experiments, we can observe that pretraining model DuET~\cite{chen2022think} fails to transfer the navigation abilities learned from simulator data to the real world. Previous zero-shot method NavGPT can follow human instructions to drive the real robot to navigate in the house. However, single-round self-thinking limits its performance on semantic-rich instructions and long instructions. Benefiting from discussions with multiple domain experts, our DiscussNav has more advanced capabilities for understanding instructions, observing fine-grained landmarks, and making decisions, which brings better performance on real robot experiments.

\subsection{Qualitative Results}
We elaborately study the DiscussNav agent's discussion and reasoning processes in the real world. As shown in Figure~\ref{fig:case}, discussions with multiple domain experts effectively facilitate DiscussNav. For the open vocabulary landmark like ``table with a robot arm'' in the given instruction, the scene observation expert (\emph{i.e.}, InstructBLIP) can only observe ``a robot sitting on top of a desk'' when the prompt is ``A scene of''  as the NavGPT~\cite{navgpt} sets. However, the scene observation expert can successfully identify the ``robotic arm'' based on the discussions between instruction analysis experts and DiscussNav. For the errors made in the previous discussion process, other experts and the DiscussNav agent have the ability to correct them in time. The second box in Figure~\ref{fig:case} displays how the landmark extraction expert corrects errors in the decomposed action sequences. When the instruction involves multiple fine-grained landmarks, the DiscussNav can first perceive the most relevant visual information through active discussion with scene observation experts and then communicate with completion estimation experts to analyze the passed fine-grained landmarks. Based on this information, the agent can conduct a more reliable chain-of-thought to make the correct movement decision. If the DiscsussNav agent feels confused about several similar movement directions when making decisions, the decision testing experts can help it recheck thought-prediction pairs based on the observation to make more suitable decisions. 

\definecolor{Gray}{gray}{0.9}
\begin{table}[ht]
\small
\centering
\caption{Real Robot Experiments in Indoor Scene.}
\label{tab:real} 
\resizebox{0.7\linewidth}{!}
{
\begin{tabular}{ll>{\columncolor{Gray}}c}
\toprule
\multicolumn{1}{c}{Training Schema} & \multicolumn{1}{c}{Method} &
\multicolumn{1}{c}{SR$\uparrow$} \\
\midrule    
\multirow{1}{*}{Pretrain + Finetune} 
& DuET~\cite{chen2022think} &0  \\
\midrule
\multirow{2}{*}{Zero-shot} 
& NavGPT~\cite{navgpt} &10  \\
& DiscussNav (Ours) &\textbf{25}  \\
\bottomrule
\end{tabular}
}
\end{table}

\section{CONCLUSION AND FUTURE WORK}
In this work, we introduce an innovative zero-shot framework for visual language navigation. We enhance the capabilities of our navigation agent, DiscussNav, by enabling it to engage in active discussions with multiple domain experts, thereby gathering valuable professional insights prior to making a move. Through our experiments, we delve into the potential of this novel paradigm in enhancing the performance of large language models on embodied tasks. Our findings underscore the benefits of discussion-based reasoning over self-thinking-based reasoning. As we look ahead, future directions include expanding the roster of large model-driven domain experts, thus facilitating richer navigation advice within discussions. Furthermore, we anticipate extending this framework to a broader range of embodied tasks by designing diverse discussion topics.

\clearpage
{
\bibliographystyle{IEEEtran}
\bibliography{IEEEabrv,reference}
}

\end{document}